\def\year{2021}\relax
\documentclass[letterpaper]{article} % DO NOT CHANGE THIS
\usepackage{aaai21}  % DO NOT CHANGE THIS
\usepackage{times}  % DO NOT CHANGE THIS
\usepackage{helvet} % DO NOT CHANGE THIS
\usepackage{courier}  % DO NOT CHANGE THIS
\usepackage[hyphens]{url}  % DO NOT CHANGE THIS
\usepackage{graphicx} % DO NOT CHANGE THIS
\usepackage{natbib}  % DO NOT CHANGE THIS AND DO NOT ADD ANY OPTIONS TO IT
\usepackage{caption} % DO NOT CHANGE THIS AND DO NOT ADD ANY OPTIONS TO IT
\usepackage{amsmath}  % MK ADDED
\DeclareMathOperator*{\argmax}{argmax}  % MK ADDED
\DeclareMathOperator*{\argmin}{argmin}  % MK ADDED
\usepackage{enumitem}  % MK ADDED
\usepackage{nidanfloat}  % MK ADDED

\nocopyright
\usepackage{subfig}

\title{Creative Captioning: An AI Grand Challenge Based on the Dixit Board Game}
\author {        Maithilee Kunda\textsuperscript{\rm 1} and
        Irina Rabkina\textsuperscript{\rm 2} \\}
\affiliations {
    \textsuperscript{\rm 1} Electrical Engineering and Computer Science, Vanderbilt University, Nashville, TN, USA \\
    \textsuperscript{\rm 2} Computer Science, Occidental College, Los Angeles, CA, USA \\
    mkunda@vanderbilt.edu, irabkina@oxy.edu
}

\begin{document}

\maketitle

%%%%%%%%%%%%%%%%%%%%%%%%%%%%%%%%%%%%%%%%%%%%%%%%%%%%%%%%%%%%%%%%%%%%
% Four sentences [Kent Beck] 
% 1.  State the problem
% 2.  Say why it’s an interesting problem
% 3.  Say what your solution achieves
% 4.  Say what follows from your solution

\begin{abstract}
We propose a new class of ``grand challenge'' AI problems that we call creative captioning---generating clever, interesting, or abstract captions for images, as well as understanding such captions.  Creative captioning draws on core AI research areas of vision, natural language processing, narrative reasoning, and social reasoning, and across all these areas, it requires sophisticated uses of common sense and cultural knowledge.  In this paper, we analyze several specific research problems that fall under creative captioning, using the popular board game Dixit as both inspiration and proposed testing ground.  We expect that Dixit could serve as an engaging and motivating benchmark for creative captioning across numerous AI research communities for the coming 1-2 decades.

%For many decades, AI has pursued board games as interesting and challenging problem areas, that inform and spur developments in many sub-areas of research, from chess to Go to poker to Hanabi.  We propose that the board game Dixit embodies many key and as yet unsolved AI research challenges, spanning computer vision, natural language understanding, social cognition, storytelling, commonsense reasoning, cultural knowledge, and creativity, to name a few.  In this paper, we discuss AI research on board games in the past; review research on Dixit from many other scientific disciplines, including psychology and education; and analyze the specific AI problems posed by Dixit, showing how solving these problems would expand the frontiers of research in many important areas.  We also discuss how a Dixit grand challenge could be structured, and how initial stepping-stone challenges could be used to catalyze research towards the larger challenge goal.
\end{abstract}

\begin{figure*}[b]
    \centering
    \includegraphics[width=\linewidth]{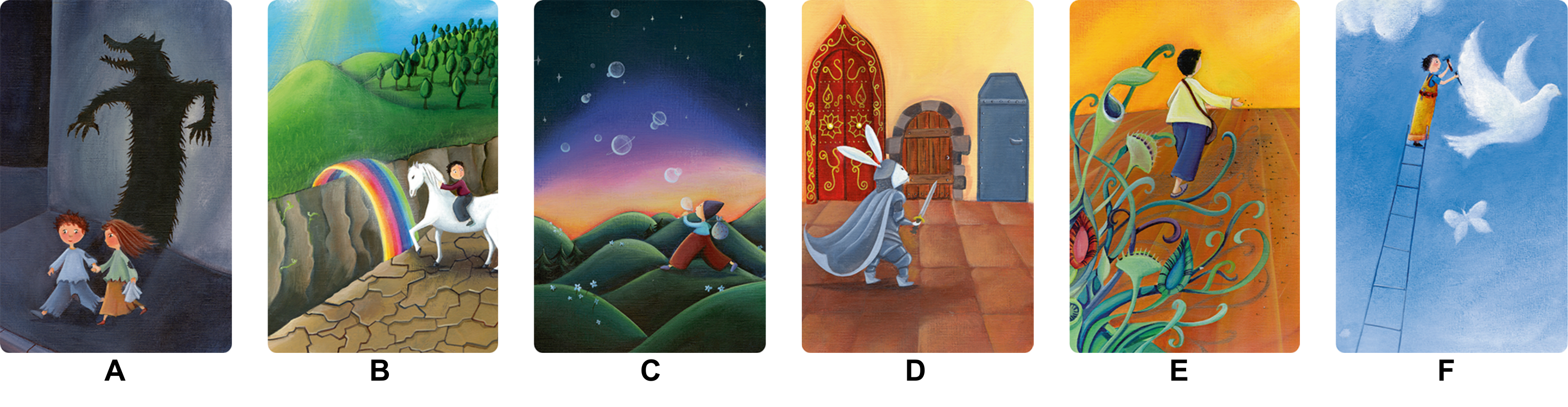}
    \caption{Images from the popular board game Dixit \cite{dixit2019resources}, which embodies several tasks in creative captioning.}
    \label{fig:dixit_cards}
\end{figure*}

%%%%%%%%%%%%%%%%%%%%%%%%%%%%%%%%%%%%%%%%%%%%%%%%%%%%%%%%%%%%%%%%%%%%
\section{Introduction}

Consider the images in Figure \ref{fig:dixit_cards}.  For each of the following phrases, which image do you think is being described?
\begin{enumerate}[]
    \item \textit{A person on a tall ladder using a hammer and chisel to make a cloud pigeon in the sky, plus a cloud butterfly.}
    \item \textit{A difficult choice among three options.}
    \item \textit{Scary.}
    \item \textit{Monty Hall.}
%    \item \textit{Michelangelo.}
    \item \textit{She got the little dog, too!}
\end{enumerate}

\newpage
Phrase 1 is fairly easy to match; it is just a literal description of the contents of Image F.  

Phrase 2 is a little bit harder; it does not use any specific nouns that refer to depicted objects, but it does refer to ``three options,'' which suggests the three doors in Image D.  Nothing in Image D directly indicates that the choice is a difficult one, or really that there is a ``choice'' at all, but we can easily imagine that the knightly rabbit is facing a choice.

Phrase 3 is harder still---``scary'' refers to an emotional quality, and so we have to consider the affect or mood conveyed by each image.  (The authors find Image A to be the most scary, though E is admittedly also creepy, especially if one has botanophobia, or fear of plants!)

Phrase 4 requires know who (or what) Monty Hall is.   Among AAAI readers, we expect that at least a few will recognize our reference to the famous Monty Hall problem of choosing a prize from behind three closed doors \cite{selvin1975on,selvin1975problem}, thus referring again to Image D.

We leave Phrase 5 as an exercise for the reader.  (Hint: This involves cultural reference to the classic American film \textit{The Wizard of Oz}, and both a song and a dialogue from it.)

We define \textbf{creative captioning} as a class of problems that includes (a) \textit{generating} clever, interesting, or abstract captions for images (as we did in creating our list of phrases), and (b) \textit{understanding} such captions (as you did in matching each phrase to an image), and related variants thereof.

Our work is inspired by the popular board game Dixit \cite{roubira2008dixit}, in which players both generate and try to match ``interesting'' captions to rather surreal-looking images, like those shown in Fig. \ref{fig:dixit_cards}.  

While Dixit provides one concrete instantiation of problems involved in creative captioning, other examples include the well-known \textit{New Yorker} cartoon captioning contest \cite{prince2017humorous,bogert2020algorithmic,li2020learning}, or generating engaging titles for inventions \cite{senda2004support} or artwork.  

Creative captioning is very related to image captioning \cite{hossain2019comprehensive}, though it goes beyond to incorporate more sophisticated aspects of vision (e.g., multiple and often abstract interpretations of an image), natural language processing (including idioms, cultural references, etc.), story or narrative reasoning (e.g., inferring, or imagining, narrative constructs related to a given image or phrase), and social reasoning (such as thinking about the intended audience for a caption).  Contributions of this paper include:

%\smallskip \noindent
%Specific contributions of this paper include:
\begin{itemize}
    \item We identify characteristics that make Dixit a fascinating window into human intelligence, by reviewing Dixit-related research in psychology and education.
    \item We describe how the Dixit board game could be set up as an AI challenge, including specific assumptions, game variants, and evaluation methods.
    \item We analyze the suite of problems that an artificial agent faces while playing Dixit, and we review the state of the art in AI research related to each problem.
\end{itemize}

%While Dixit provides one concrete instantiation, other problems in creative captioning include, for example, the well-known \textit{New Yorker} cartoon captioning contest , or methods for generating good titles for things \cite{senda2004support,Pallan2016AutomaticallyAT}. 

%bachelors thesis on deep learning something for new yorker captioning contest \cite{li2020learning}

%mentions captioning political cartoons as a creative AI task (or something) \cite{bogert2020algorithmic}

%%%%%%%%%%%%%%%%%%%%%%%%%%%%%%%%%%%%%%%%%%%%%%%%%%%%%%%%%%%%%%%%%%%%

%... In this paper, we propose the Dixit game as a challenge problem for creative captioning. 

% \section{Board Games in AI}

% chess
% go

% these games involve a giant search tree, and a lot of the AI advances were about how to search.  Dixit is not like that.

% - side note: infinite states?

% poker involves social reasoning?  (does it?)

% hanabi (google challenge)

%%%%%%%%%%%%%%%%%%%%%%%%%%%%%%%%%%%%%%%%%%%%%%%%%%%%%%%%%%%%%%%%%%%%
\section{The Dixit Game}
\begin{figure*} [t]
    \centering
    \includegraphics[width=\linewidth] {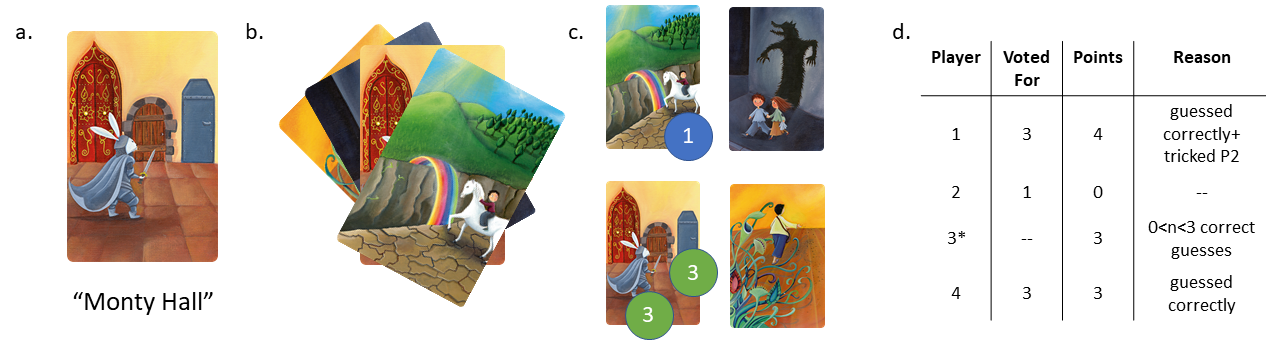}
    \caption{An example of a round of Dixit. (a) The storyteller selects a card and utters a corresponding word/phrase. (b) Other players select a card from their hand and play it. (c) All cards are turned face up and non-storyteller players vote on the card they believe was played by the storyteller. (d) Finally, points for the round are calculated. In the points table in (d), we mark the storyteller, player 3, with an *. All images of cards were retrieved from the \citet{dixit2019resources} publisher website.}
    \label{fig:dixit-game}
\end{figure*}

Dixit \cite{roubira2008dixit} is a tabletop card game typically played with 4-6 players. The game contains 84 cards, 36 numbered tokens, 6 player markers, and a game board. Cards are storybook-style illustrations, often surreal (Figure \ref{fig:dixit_cards}).  
Game play takes place as follows (Figure \ref{fig:dixit-game}).

% \begin{figure} [h]
%     \centering
%     \subfloat{
%     {\includegraphics[width=.4\linewidth] {images/DIXIT_CARD_6_72dpi}
%     }}
%     \subfloat{
 
%     {\includegraphics[width=.4\linewidth]{images/DIXIT_CARD_4_72dpi}}}

%     \caption{Two example Dixit cards.}
%     \label{fig:dixit-cards}
% \end{figure}

\emph{Setup.}
%At the beginning of the game, 
Each player %selects a player marker and places it on the spot marked "0" on the game board, representing the fact that each player 
starts with zero points. The deck is shuffled, and each player is dealt six cards. Each player also receives %some number of 
tokens to be used for voting.  %, corresponding to the number of players in the game (i.e., in a four-player game, each player would receive four tokes, numbered 1-4). 
A player is selected at random to be the first storyteller.

\emph{Storyteller Turn.}
At the beginning of each round, the storyteller secretly selects a card from their hand. They then utter a clue or label for the card, and place their card face down on the table.  Per the Dixit instructions: 
\begin{quote}
    ``[This] clue can take many different forms: it can be made up of one or more words or can even be a sound or group of sounds that represent the clue. It can be invented on the spot or it can take the form of already existing works (a part of a poem or song, a movie title, a proverb, etc...).'' \cite{roubira2008dixit}
\end{quote}

\emph{Remaining Players' Turn.}
Then, each remaining player picks a ``decoy'' card from their own hand that they associate with the storyteller's utterance. These cards are placed face down on top of the storyteller's card.

\emph{Voting.}
After all players have added their cards to the pile, the storyteller shuffles and then reveals all of the cards. %The cards are then turned over one by one, and assigned a number (typically, the card on the far left is \#1, the next card is \#2, etc.). 
Each player, other than the storyteller, then votes for the card that they believe to be the storyteller's by placing the corresponding token face down in front of them.  (Tokens are placed face-down so that votes are hidden while players are still making decisions.) Players cannot vote for their own card.

\emph{Scoring.}
Each round's point distribution is determined by the players' votes. If \textit{all} players or \textit{no} players correctly guessed the storyteller's card, the storyteller does not earn any points, and all other players earn 2 points. Otherwise, the storyteller earns 3 points, as does each player that guessed correctly. Each player, other than the storyteller, earns 1 additional point for each vote that their own card received. %person who incorrectly voted for their card. Players move their player marker forward, according to the number of points they earned.

\emph{Next Turn.}
All players draw from the deck to replenish their hand back up to six cards. The storyteller role is passed clockwise.  %, to the player on the old storyteller's left.

\emph{Game ending.}
The game ends once any player has reached 30 points or when the last card has been drawn from the deck, at which time the player with the most points wins.

\emph{Dixit as creative captioning.}
Dixit involves multiple forms of creative captioning, including when the storyteller chooses a card and generates a clue/phrase for it; when the other players select their own decoy cards; and when players vote on the card they think is the original storyteller's card.

% game citation \cite{roubira2008dixit}

% total 84 cards

% each person gets 6 cards.

% dealer picks one card, and without showing it to others, one phrase story or description (one or a few words, or even a sound!)
% all cultural references are on the table

% point: playing with friends would imply that one is also drawing on personal knowledge of that person.  to make this ``fair'' for an AI agent, we could constrain this to be playing with strangers only.  assume cultural knowledge?  or something?

% point: there may be table chatter in games with people.  this would complicate things considerably, so we exclude this for our AI benchmark

% if everyone guesses the right card, or no one, then dealer gets zero points and everyone else gets 2

% in all other cases, dealer gets 3 points (lump sum)
% everyone else who guesses dealer's card gets 3 points
% if people guess someone else's distracter card, then the distracter card person gets 1 point

%%%%%%%%%%%%%%%%%%%%%%%%%%%%%%%%%%%%%%%%%%%%%%%%%%%%%%%%%%%%%%%%%%%%
\section{Research on Dixit with People}

Dixit (or Dixit-inspired activities) represents a fascinating task format for eliciting human creative captioning, as evidenced by its widespread use in a variety of research studies across different fields.  Table \ref{fig:papertable} lists a sampling of these studies.  For example, one study used a card game similar to Dixit as a research tool for querying people's cultural knowledge \cite{bekesas2018cosmocult}.  

In another quite clever deployment of the game, Dixit was used to teach software design patterns to graduate CS students.  Essentially, students played Dixit as usual except that their ``clues'' had to be drawn from the vocabulary of programming design patterns taught during the course.  As an example from this study \cite{piccolo2010using}:

\begin{quote}
    ``In his turn, a player select in his hand a card that has a humanized rabbit looking for three different doors. He thinks that this card relates to the Strategy pattern, where you can choose different implementations for an algorithm. Then, he put his card backwards in the table, saying "Strategy". Each other player them should select the card in his hand that he thinks that is the most related to the Startegy pattern. For instance, another player should select a card with several water drops, relating that in the Strategy pattern there are several classes encapsulated in the same abstraction.''
\end{quote}

%% IN GOOGLE SHEETS TABLE

%teach software design patterns using Dixit \cite{piccolo2010using}

%trying to teach people about art using these games \cite{vayanou2019play}

%taxonomy for board games.  Dixit has low rule complexity \cite{chircop2016experiential}

%using new card game (inspired loosely by Dixit and other games) where the new game is a research tool for querying people's cultural knowledge and use of that knowledge?) in young people in brazil \cite{bekesas2018cosmocult}

%using Dixit to study group communication \cite{vitancol2018dixit}

%looking at patient population, how they choose symbols and metaphorical language and stuff \cite{mousnier2016dixit}

% something (possibly not actually related) about some human computation thing inspired by dixit \cite{musat2013novel}

\begin{table*}
    \centering
    \caption{Sampling of research studies using Dixit from psychology, education, and other social and cognitive science areas.}
    \label{fig:papertable}
    \includegraphics[width=\linewidth]{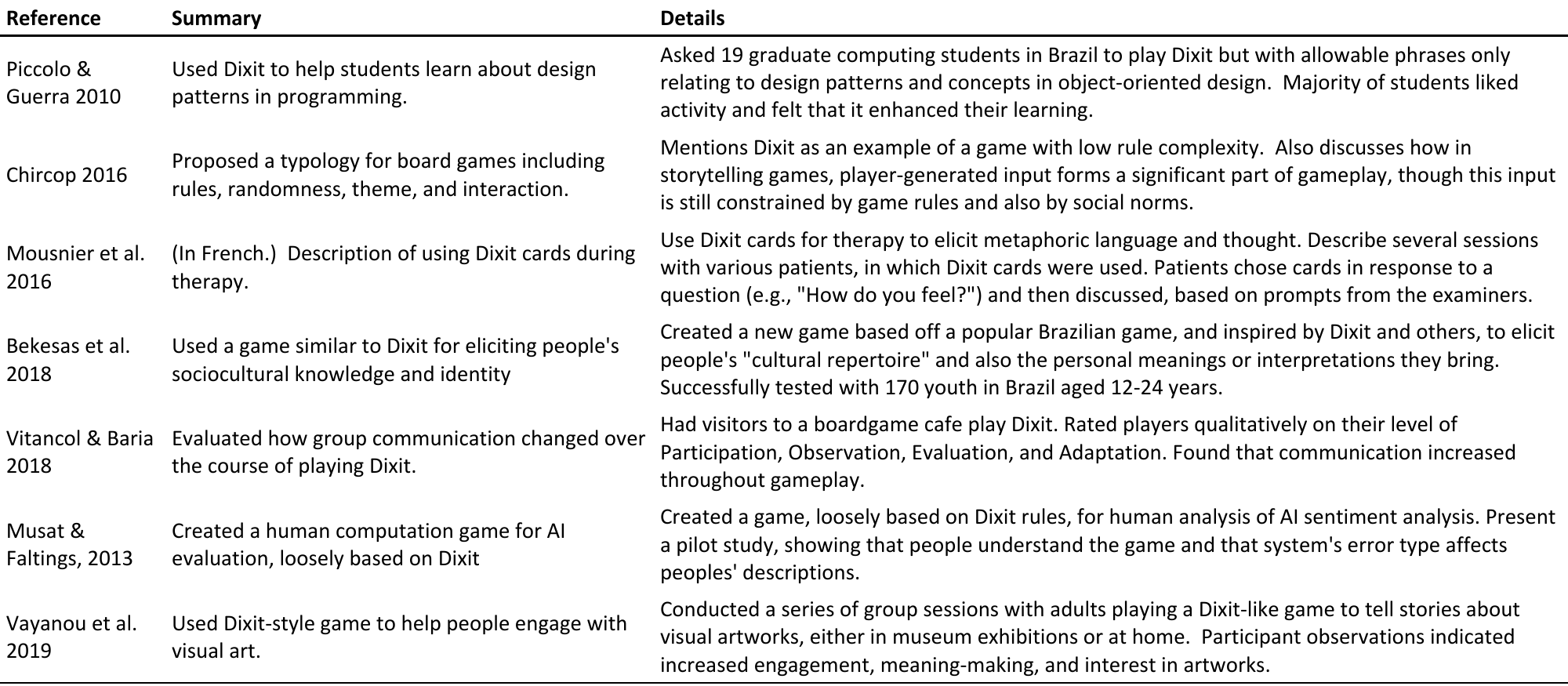}
\end{table*}

%% MK IN PROGRESS

%% IR IN PROGRESS

%% HAVE PDF BUT NOT YET IN TABLE

Other studies include using Dixit as part of board game training to make people smarter \cite{bartolucci2019board}; analyzing Dixit within a taxonomy of narrative board and card games \cite{sullivan2017taxonomy}; discussing Dixit as an adaptation of Rorschach tests \cite{rogerson2017f}; using Dixit cards to spur ideation for game design \cite{wetzel2017developing}; using Dixit cards as a method for language sampling in children that elicits more lexical diversity than traditional methods % - elicits more lexical diversity ** good sign for using this as an AI task, is difficult!
\cite{smith2018dixit}; Dixit as edutainment \cite{novikova2015smart}; how Dixit can be used to teach complex concepts like ethics \cite{mazurkiewicz2013dixit}; discussions of the shared narrative experience among players of games like Dixit \cite{montanarini2019space}; and using Dixit for therapy \cite{ikiz2020dixit}.

%% NO PDF AND NOT YET IN GOOGLE SHEETS TABLE

%something about making a game with a bunch of cards?  in korean
%\cite{ahn2016model}

%%%%%%%%%%%%%%%%%%%%%%%%%%%%%%%%%%%%%%%%%%%%%%%%%%%%%%%%%%%%%%%%%%%%
\section{Dixit AI Challenge}
While the rules of Dixit are easy to follow, we believe that the game will pose a significant challenge to modern AI agents. As such, we propose Dixit as a creative captioning AI challenge under the following parameters.  (We use the abbreviation DA to refer to a Dixit Agent.)

We note that there are two ways to evaluate a DA: winning the game and playing it in a believably human-like way. However, omitting the latter requirement changes the game considerably, in that human players would then be voting and acting based on their own mental models of how they think an artificial agent would be playing the game.  Thus, our proposed Dixit AI challenge assumes that human players do not know which is the artificial agent (or perhaps that there is an artificial agent at all).  %we believe that the former requires the latter---and that either can be used to evaluate creative captioning---so we do not distinguish between these evaluation criteria here.
% Question:  Should the DA have to win, or pass the Turing Test, or both?  (Is one necessary for the other?)  Idea:  Turing Test is a prerequisite for this challenge.  If the human players know that the DA is a computer, then THEY might be playing/voting according to their own mental models of how a computer ought to be playing the game.  So to play Dixit ``as intended,'' we say that the human players do not know that the DA is a computer... (perhaps they don't even know that a computer is playing at all!) which essentially requires that the DA is able to give human-passable responses.

\begin{enumerate}
    \item The DA must be able to play a full Dixit game against human players. That is, it must be able to play both storyteller and non-storyteller roles.
    \item The basic Dixit game is intended for 4-6 human players, and thus the DA must be able to play against anywhere from 3-5 human players.
    \item While the official rules of Dixit allow for game phrases to include ``noises'' or, in some variants, physical gestures or charades, the proposed AI challenge will allow only text-based phrases.
    \item Dixit calls for game phrases to be ``short.''  We propose imposing a hard limit $K$ on phrase length as a game parameter. For example $K=4$ would be quite reasonable.
    \item The game will be played over a virtual connection, such that the only communications among players consists of selections of cards, plain-text phrases, and votes.  Thus, we eliminate the roles of facial expressions, body language, prosody, and other forms of nonverbal communication.  (Perhaps we leave these for the next AI Dixit challenge!)
    \item Table chatter is not permitted amongst players, i.e., there is no extraneous conversation allowed.
    \item The game will take place using a specified language (e.g., English), cultural context (e.g., the United States), and player characteristics (e.g., adults, or college students, or 10-year-olds, or college computer science students, etc.).  We expect that DAs will be designed initially for one specific set of these characteristics, but the core AI methods developed ought to (eventually) be able to generalize across these different game contexts.
    \item The human players should be strangers to each other; this ensures that the players are not relying on personal knowledge, inside jokes, etc., that would be impossible for the DA to know or understand.  However, the DA and human players may well observe and learn from each other's behaviors during the course of the game.
    \item The game will be played with previously unseen cards (e.g., from an expansion pack) that neither the DA nor any of the players have previously seen.  
    \item The DA must be able to explain all of its actions (i.e., why it selected a phrase or card). This is to prevent winning by way of Eliza effect \cite{o1984eliza}, and is a reasonable requirement to apply to human players as well.
\end{enumerate}

We intend for these parameters to limit the difficulty of the game, while maintaining its spirit. For example, while table chatter adds entertainment for human players and may often affect the course of game play, we remove it as a simplifying assumption for our initial Dixit AI challenge.   Explaining one's choices, however, is common among human players \cite{piccolo2010using,vayanou2019play} and serves to show that choices were not made completely at random.

\section{Creative Captioning Problems in Dixit}

Winning a full game of Dixit involves maximizing one's score earned across both storyteller and non-storyteller rounds.  Here, we formalize specific problems and subproblems that a Dixit Agent (DA) would have to solve in order to win a game against human players. %  We then decompose these problems into technical subproblems.  For each subproblem, we provide a sampling of relevant  AI literature and argue for why the subproblem is an interesting open challenge.  
Of course, there are many other possible ways to frame the problems that Dixit poses; what we present here is one possible starting point.

\subsection{Storyteller round}

In the storyteller round, the DA begins with a hand $H$ of six cards, each represented as a single color image: $H = \{C_1, ..., C_6\}$.

Then, the DA must select one card $C_{target}$ and produce a corresponding text-based phrase $X_{target}$ that goes with that card.  The phrase can be anything from the space of all possible utterances $\mathcal{U}_k$ of length $k$ in the language in which the game is being played, i.e., $X_{target} \in \mathcal{U}_k$.

That is actually all that is required from the DA during the storytelling round.  The rest of the round depends entirely on how the other players react.

So how does the DA produce a ``winning'' choice of $C_{target}$ and $X_{target}$?  This is actually a somewhat bizarre and ambiguous optimization problem.

Let $n \in {4, 5, 6}$ be the total number of players in the game.  Then, if the DA is the storyteller, there are $n-1$ voting players in that round.

The score that the DA will receive as the storyteller depends on the number of players $n_V$ that vote for its card:
\begin{equation*}
    score =\begin{cases} 
      0 & n_V = 0 \\
      3 & 0 < n_V < n-1 \\
      0 & n_V = n-1
   \end{cases}
\end{equation*}
   
For any choice of card $C_i$ and phrase $X_i$, the DA must essentially estimate the probability that the number of voting players $n_V$ will lie in the desired range.  If $X_{target}$ is too specific to $C_{target}$, then it is likely that $n_V = n-1$.  If $X_{target}$ is not specific enough to $C_{target}$, then it is likely that $n_V = 0$.

Given the ability to estimate this probability $P_{scoring}$, the DA is then trying to choose some combination of $C_{target}$ from its hand $H$, and $X_{target}$ from the set of all possible utterances $\mathcal{U}_k$ of length $k$, that maximizes $P_{scoring}$.  We call this \emph{Storyteller Strategy \#1}:
\begin{equation} \label{eq:storyteller_probability}
P_{scoring}(C_i, X_i) = P(0 < n_V < n-1 \mid 
C_i, X_i) 
\end{equation}
\begin{equation} \label{eq:storyteller_probstrategy}
\begin{bmatrix}C_{target} \\ X_{target}\end{bmatrix} = 
\argmax_{C_i \in H}{\left\{
\argmax_{X_j \in \mathcal{U}_k}{(P_{scoring})}
\right\}}
\end{equation}

However, \textit{other} players can also earn points during this round, based on whether they vote for the storyteller's card (3 points), and also whether other players vote for their card (1 point per other player that has been deceived).  So, while acting according to Eq. \ref{eq:storyteller_probstrategy} can maximize the chances that the DA will earn 3 points, the DA's \textit{net} lead over the other players will vary, depending on how many other players vote for the storyteller's card.  

The net points earned by other players will be minimized if only one player votes for the DA's card.  Thus, a different strategy for the DA during storytellers round is to minimize the expected number of players $E_{votes} = E[n_V]$ who might vote for the DAs card, while keeping it above 0.  We call this \textit{Storyteller Strategy \#2}.
\begin{equation} \label{eq:storyteller_expectedvotes}
E_{votes}(C_i, X_i) = E[n_V \mid 
C_i, X_i]
\end{equation}
\begin{equation} \label{eq:storyteller_expectstrategy}
\begin{bmatrix}C_{target} \\ X_{target}\end{bmatrix} =
\argmin_{C_i \in H}{\left\{
\argmin_{X_j \in \mathcal{U}_k}{\left(E_{votes} \mid E_{votes}>0\right)}
\right\}}
\end{equation}

\subsubsection{How Many Votes subproblem.}
Either of the above strategies can be roughly decomposed into two subproblems.  First, given any candidate pairing of a card and a phrase $[C_i, X_i$, the DA needs to be able to estimate how many other players are likely to vote for it, as in Eq. \ref{eq:storyteller_probability} or Eq. \ref{eq:storyteller_expectedvotes}.  As mentioned above, we require that the deck of cards is not previously known to the DA before gameplay, and so these probabilities cannot be ``precomputed'' for given cards and pre-selected target phrases.

Solving this subproblem requires several different technical AI capabilities, including:
\begin{enumerate}[nolistsep,noitemsep]
    \item \textit{Vision:} What does the card $C_i$ actually depict?  What are the objects, characters, scene information, and affective and/or cultural implications?
    \item \textit{NLP:} What does the phrase $X_i$ actually mean?  What is the proper parsing, word or phrase meanings, and affective and/or cultural implications?
    \item \textit{Story reasoning:} Given the card and phrase pairing, how can they be interpreted together to form a coherent visual+linguistic story?
    \item \textit{Social reasoning:} Given the card and phrase pairing, how many other players are likely to vote for this card, also relative to the other potential decoy cards that other players might produce in response to the phrase $X_i$?
\end{enumerate}

\subsubsection{Find a Phrase subproblem.}  Eq. \ref{eq:storyteller_probstrategy} and Eq. \ref{eq:storyteller_expectstrategy} require the DA to choose a target card $C_{target} \in H$ and a target phrase $X_{target} \in \mathcal{U}_k$ from all possible utterances of length $k$.

One way to solve this subproblem could be to iterate through all $C_i$ in the DA's hand $H$, which only has six cards, and then for each $C_i$, have some search strategy that generates a series of candidate phrases $X_i$ and evaluates each one according to the How Many Votes subproblem.

Then, the core remaining subproblem becomes how to generate a series of candidate phrases $X_i$ for a given $C_i$.  Solving this subproblem requires:

\begin{enumerate}[nolistsep,noitemsep]
    \item \textit{Vision:} As above.
    \item \textit{NLP and story reasoning:} What are possible ``creative story captions'' that could be applied to describe $C_i$?
\end{enumerate}

\subsection{Non-storyteller round}

When the DA is not the storyteller, the round begins when the storyteller (another player) produces a target phrase $X_{target}$ for that round.  In these rounds, the DA has two somewhat separable goals.

First, the DA must choose a card $C_{decoy} \in H$ from its hand that best lures other players into voting for it.  During the voting phase, the DA will receive 1 point for each player that votes for its card.  Thus, the DA should choose a card from its hand that maximizes the number of players likely to vote for it.  As above, we let $n_V$ denote the number of other players voting for the DA's card:
\begin{equation} \label{eq:nonstoryteller_cardstrategy}
C_{decoy} =
\argmax_{C_i \in H}{\left\{
E_{votes}\mid C_i, X_{target}
\right\}}
\end{equation}

Second, the DA will see a set $S$ of cards $C_i$ on the table (one from the storyteller, one from itself $C_{decoy}$, and one from each other player), and it must vote for the card $C_{vote}$ that it thinks is the storyteller's card:
\begin{equation} \label{eq:nonstoryteller_probability}
P_{target}(C_i) = P(C_i = C_{target} \mid X_{target})
\end{equation}
\begin{equation} \label{eq:nonstoryteller_vote}
C_{vote} =
\argmax_{C_i \in S}{\left\{
P_{target}(C_i)
\right\}}
\end{equation}

\subsubsection{How Many Votes subproblem variants.}
In the non-storyteller round, when the DA is selecting its decoy card, it is solving something very similar to the How Many Votes subproblem as described above, except now it is in the DA's interest to get as many players as possible to vote for its card.  And, because the DA is searching its hand $H$, there are only six possible cards $C_i$ to choose from.  Thus, a simple approach would be for the DA to iterate through all six cards in its hand, compute the expected number of votes for each, and select the card giving the maximum estimate.

Of course, as noted above, solving the How Many Votes subproblem is quite difficult and requires vision, NLP, story reasoning, and social reasoning.

Finally, when the Dixit agent has to vote on the card that it believes is the original storyteller's card, it is again solving a variant of the How Many Votes subproblem.  It can perform exhaustive search through the $n-1$ available card options (one card from each player except its own), and for each one, estimate the probability of its being the storyteller's card.

\subsection{End-Game Considerations}

The above equations describe several strategies that the DA can use to essentially maximize its own score.  However, there are also situations in Dixit when the DA might need to instead shift strategies to prevent other players from scoring, for instance towards the end of a game if another player is very close to winning.

For example, suppose the DA is not the storyteller, and the player who is the storyteller is within 3 points of winning.  Then, instead of using Eq. \ref{eq:nonstoryteller_vote} to choose its vote, which maximizes the probability in Eq. \ref{eq:nonstoryteller_probability}, the DA might instead want to minimize this Eq. \ref{eq:nonstoryteller_probability} probability, in order to prevent the storyteller from winning.

Many games have strategies that shift as gameplay progresses.  In Dixit, all players can see the scoreboard at all times, and so while reasoning about the scores of other players might not always be strictly necessary to win, it does play a potentially useful (and potentially game-changing) role.

%%%%%%%%%%%%%%%%%%%%%%%%%%%%%%%%%%%%%%%%%%%%%%%%%%%%%%%%%%%%%%%%%%%%
\section{Towards Creative Captioning: The Current State of the Art}

Creative captioning in general, and in particular the specific challenge we propose of winning a game of Dixit, touches on core problems for many subfields of AI, including (1) vision, (2) natural language processing, (3) story or narrative reasoning, and (4) social reasoning, among others. In this section, we discuss the current state of the art in each of these subfields individually and taken as an integrated whole.  %and suggest how these may be unified for creative captioning in a Dixit Agent (DA).

\subsection{Vision}
Creative captioning requires several robust vision capabilities, as described (non-exhaustively) below.
%Creative captioning is, first and foremost, a captioning task. Thus, reasonable vision-related prerequisites include the following.

\subsubsection{Object recognition: What is in the image?}  
The last eight years have seen a revolution in approaches to object recognition, due in part to advances in dataset size \cite{deng2009imagenet} and deep learning methods \cite{krizhevsky2012imagenet}.  However, generalized object recognition is still a difficult problem, for example when models are faced with new images that are more complex than training images \cite{recht2019imagenet} or that depict objects in unusual poses \cite{barbu2019objectnet} or sociocultural contexts \cite{de2019does}.  Additional challenges emerge when considering not just photographic images but also artwork and other other visual styles of depiction \cite{hall2015cross,westlake2016detecting}.  Dixit game images in particular, as shown in Figure \ref{fig:dixit_cards} are especially challenging for artificial vision systems because they include surreal elements \cite{florea2016recognizing}.

%art interpretation / surrealist / cartoony

%comic book paper

%Understanding the objects/entities in an image  
%in order to be successful in creative captioning, a system must be able to perceive and interpret a visual image. This may require scene analysis and object identification, along with more abstract reasoning, like interpretation of the affective properties of an image.

\subsubsection{Scene analysis: How are the objects related?}  
In addition to identifying objects and entities in an image, creative captioning additionally requires understanding the scene, i.e., relationships among objects.  Going beyond just identifying objects and their relationships \cite{dai2017detecting}, however, creative captioning also requires inferring aspects of the relevance of the scene to common scenes, common sense interpretations such as inferring physical interactions among objects \cite{battaglia2013simulation}, cultural contexts, etc.  

\subsubsection{Affective analysis: What is the emotional content of an image?}  
Creative captioning also requires inferring affective aspects of an image, including overall mood or tone \cite{machajdik2010affective}, facial expressions of characters \cite{zhao2016review}, etc.

%mood of image

%facial expressions of characters

%affective/emotional properties of imag

%rat in middle-eastern/south asian garb (with minareted building in background) playing a snake charming instrument thingy that is itself a snake

%many are ambiguous

%like black splotch monster thing... what IS that?  hard even for a person to describe in words! (Rorschach?)

%can't label everything.... can't label certain objects, can't label ALL relationships, can't label social context

%need: fully searchable visual index of everything you've seen!

%%%%%%%%%%%%%%%%%%%%%%%%%%%%%%%%%%%%%%%%%%%%%%%%%%%%%%%%%%%%%%%%%%%%
\subsection{Natural Language Processing/Understanding}
 In recent years, language models such as Bert \cite{devlin2018bert}, GPT \cite{radford2018improving}, and their derivatives \cite[etc.]{liu2019roberta, radford2019language, brown2020language} have made substantial progress on many natural language tasks, ranging from question answering to story completion. Of these, story completion (e.g., HellaSwag \cite{zellers2019hellaswag} and StoryCloze \cite{mostafazadeh2016corpus}) is the most similar to creative captioning.
 
 Story completion tasks, sometimes referred to as cloze tasks, combine language understanding with language generation. The datasets contain short (3-5 sentence length) stories, the first few sentences of which are provided as input to the system being tested. The system must then generate the rest of the story. This requires understanding the contents of the story, along with any implied commonsense reasoning and storyteller intentions. Humans perform incredibly well on these tasks---100\% accuracy on StoryCloze \cite{mostafazadeh2016corpus} and 95.6\% accuracy on HellaSwag \cite{zellers2019hellaswag}. Yet, even the newest (at the time of writing) GPT model, GPT-3, performs significantly worse \cite{brown2020language}. This is likely because language models do not perform natural language understanding or commonsense reasoning, and instead find patterns and make connections across millions of training examples \cite{marcus_davis_2020}. We believe that this will cause language models to struggle on the creative captioning task as well, especially since generation must occur between modalities (i.e., generate language based on an image, or select an image based on language). On the other hand, language models have been successful on some creative tasks (e.g., poetry \cite{liao2019gpt}), which suggests that creative captioning may not be entirely out of reach.
 
 Other approaches combine knowledge and inference for language understanding. For example, \cite{lin2017reasoning} encoded three types of commonsense knowledge as inference rules: event narrative knowledge, entity semantic knowledge, and sentiment coherence knowledge. They then learned an attention model which selected appropriate rules for a given question. This inference-based model outperformed several others, including a Deep Structured Semantic Model \cite{mostafazadeh2016corpus} and an LSTM-based Recurrent Neural Network \cite{pichotta2016learning}, on a version of the StoryCloze task. Others \cite{botschen2018frame} have found that incorporating knowledge from sources like FrameNet \cite{baker1998berkeley} and Wikidata \cite{vrandevcic2014wikidata} similarly improve performance on another cloze task \cite{habernal2017argument}. While, to the best of our knowledge, such knowledge based approaches have not been tested on creative tasks, their strong performance on cloze tasks suggests that they may be able to handle the language interpretation component of creative captioning, as well.

%this maybe is related to the various stages of image<->phrase mappings

%%%%%%%%%%%%%%%%%%%%%%%%%%%%%%%%%%%%%%the%%%%%%%%%%%%%%%%%%%%%%%%%%%%%%
\subsection{Story Reasoning/Narrative Understanding}

The ``creative'' part of creative captioning goes beyond traditional image captioning tasks, and their emphasis on veridical image description, to include more sophisticated interpretations of images and phrases.  A creative captioning agent must be able to generate multiple alternatives when interpreting images, phrases, or both; and also to consider such interpretations at multiple levels of abstraction.

This class of capabilities is strongly tied to AI research in story reasoning / narrative understanding, as human interpretations of images and phrases often revolve around story-like conceptual constructs, like the ``Monty Hall'' example in Figure \ref{fig:dixit_cards}.  

AI research on narrative reasoning observes that such reasoning often relies on an agent's prior knowledge base of stories or story prototypes, as well as rich analogical reasoning to build or reason about new stories \cite{finlayson2009deriving}.  There are many open research questions in modeling story structures \cite{riedl2006linear} as well as how to elicit data for training story reasoning systems \cite{li2013story}.

Moreover, creative captioning exemplifies narrative reasoning that bridges both linguistic and visual inputs, requiring at some level unified conceptual representations that can span both modalities.  While much AI research in narrative reasoning has relied on linguistic representations, there is also work on such reasoning in images \cite{cohn2020your,iyyer2017amazing}.

%image captioning

%storytelling / narrative reasoning 

%no particular role for prosody

%special note: the instructions say that people can also just make a ``sound'' so ideally should have sound processing also.  hum a bit of a song?

%%%%%%%%%%%%%%%%%%%%%%%%%%%%%%%%%%%%%%%%%%%%%%%%%%%%%%%%%%%%%%%%%%%%
\subsection{Social Reasoning}
Creative captioning must be creative, but not so much so that it is uninterpretable; people must be able to understand the reference. This requires sufficient social reasoning to consider what connections others are likely to make, what cultural references they are likely to be aware of, and how they are likely to approach creative captioning more broadly. In the context of the Dixit game, the Dixit Agent (DA) must also reason about the strategies other players are likely to pursue.

The most similar social reasoning problems to those posed by creative captioning and the Dixit game are in other game domains. For example, Hanabi and Werewolf have both been proposed as AI challenges \cite[respectively]{bard2020hanabi, toriumi2016ai} specifically because they require social reasoning for successful gameplay. In this section, we focus our discussion on these two games.

Hanabi is a cooperative card game, in which players work to construct ordered decks of cards (1-5) according to their colors (white, yellow, blue, green, red). Players are limited in both information and in communication: players cannot see their own cards, they have a limited number of hints to give each other, and those hints can only contain information about card color or number---not both. To succeed, players must consider not only the information explicitly given in a hint, but also the information implied by it. For example, if the red 1 has just been played, and a player is then given the hint that they have a red card, they might infer that the card is, in fact, a playable red 2.

\citet{bard2020hanabi} present several baseline agents for the Hanabi challenge. These include both rule-based and learning agents. In a self-play setting, the rule-based agents outperform the learning agents. However, only the learning agents are tested in ad-hoc teams (i.e., teams of different agent types) because of the rigidity built into the rule-based agents. None of the agents were tested while playing on mixed teams with humans, and none explicitly took social reasoning into account. Yet, \citet{eger2020operationalizing} found that giving agents the ability to reason about the intents of their human teammates led to improved scores on mixed human-AI teams. Similarly, \citet{liang2019implicit} found that human Hanabi players are more likely to believe they are playing with other people when AI agents perform explicit social reasoning (in this case by considering the possible interpretations of implied information communicated by hints). This is even more likely to be true of Dixit, where metaphorical interpretation of communication is important to game play.

Unlike Hanabi, Werewolf is a competitive game in which winning strategies require some level of deception. At the beginning of gameplay, players are privately assigned roles, corresponding to one of two teams: townspeople and werewolves\footnote{Some roles have special abilities that, for clarity, we do not discuss here.}. While each werewolf knows who the other werewolves are, the townspeople do not know any other player's role.

Each round is separated into a day phase and a night phase. During the day, there is open discussion. Townspeople strategize and attempt to discern who the werewolves are. Werewolves, on the other hand, try to throw the townspeople off their trail. At the end of the day phase, players vote for who they believe to be a werewolf and that person is removed from the game. At night, the werewolves select a townsperson, a victim, to also remove from the game. If there are more werewolves than townspeople at any point, the werewolves win. If however, all werewolves are voted out, the townspeople win.

The biggest challenge of the Werewolf game is the open conversation during the day, which requires complete conversational AI. To limit this complexity, \citet{toriumi2016ai} limit both the number and type of utterances allowed for players in their challenge. Nonetheless, most approaches to Werewolf-playing agents base behavior on the game logs of human players, including transcription of the conversations between them \cite[etc.]{hirata2016werewolf, hancock2017towards, kondoh2018development, shoji2019strategies}. These conversations encode the speakers' social reasoning, often explicitly by lying or calling out presumed lies. We believe that a similar approach (i.e., learning from observation of human players) may be fitting for a DA, as well.

%%%%%%%%%%%%%%%%%%%%%%%%%%%%%%%%%%%%%%%%%%%%%%%%%%%%%%%%%%%%%%%%%%%%
\subsection{Putting it All Together: Integrated Reasoning}
A successful DA needs to have strong abilities in each of the subareas described above. Perhaps more importantly, however, it needs to integrate these abilities into a unified system. This requires being able to not only reason about multiple modalities (i.e., images and natural language) but also to unify multiple reasoning styles (i.e., story understanding and social reasoning).

The systems that are closest to such integration are cognitive architectures \cite[etc.]{anderson2005human, langley2006unified, laird2012soar, forbus2017analogy}. Because they are designed as a single system that performs multiple types of reasoning over multiple modalities, they are well-positioned for a task like creative captioning. However, the state of the art in most most of the subproblems needed for creative captioning (i.e., visual perception, natural language processing, etc.) has been set by deep learning systems, rather than cognitive architectures. The team behind Watson found that a combination of statistical and knowledge-based approaches was necessary to beat humans at Jeopardy! \cite{ferrucci2010building}. Perhaps integrating approaches from deep learning and cognitive architecture will similarly lead to beating humans at Dixit, as well success in creative captioning more broadly.

%%%%%%%%%%%%%%%%%%%%%%%%%%%%%%%%%%%%%%%%%%%%%%%%%%%%%%%%%%%%%%%%%%%%
\section{Conclusion}
We have identified \emph{creative captioning} as a novel challenge for AI systems. To be successful at creative captioning, we argue that an agent must, at the very least, integrate visual perception, natural language understanding, and social reasoning. To that end, we have proposed the game Dixit as a domain for creative captioning, and identified intermediate subproblems that must be solved along the path to both successful play in Dixit and creative captioning overall. In the future, we will work toward solving these subproblems. We hope our colleagues will join us.

\bibliography{references.bib}
\end{document}